\documentclass{article}

\PassOptionsToPackage{numbers, compress}{natbib}

\usepackage[final]{neurips_2021}
\usepackage[utf8]{inputenc} 
\usepackage[T1]{fontenc}    
\usepackage{hyperref}       
\usepackage{url}            
\usepackage{booktabs}       
\usepackage{amsfonts}       
\usepackage{nicefrac}       
\usepackage{microtype}      
\usepackage{xcolor}         
\usepackage[normalem]{ulem} 
\usepackage{chemformula}
\usepackage{subfig}
\usepackage{amssymb}
\usepackage{mathtools}
\usepackage[c2]{optidef}
\usepackage{wrapfig}
\usepackage{siunitx}
\usepackage{caption}

\title{Coalitional Bargaining via Reinforcement Learning: An Application to Collaborative Vehicle Routing}

\author{%
  Stephen Mak \\
  University of Cambridge \\
  \And
  Liming Xu \\
  University of Cambridge \\
  \And
  Tim Pearce \\
  Tsinghua University \\
  \And
  Michael Ostroumov \\
  Value Chain Lab \\
  \And
  Alexandra Brintrup \\
  University of Cambridge \\
}

\begin{document}
\renewcommand{\sectionautorefname}{Section}
\renewcommand{\subsectionautorefname}{Subsection}
\maketitle

\begin{abstract}
  \label{sec: abstract}
  Collaborative Vehicle Routing is where delivery companies cooperate by sharing their delivery information and performing delivery requests on behalf of each other. This achieves economies of scale and thus reduces cost, greenhouse gas emissions, and road congestion. But which company should partner with whom, and how much should each company be compensated? Traditional game theoretic solution concepts, such as the Shapley value or nucleolus, are difficult to calculate for the real-world problem of Collaborative Vehicle Routing due to the characteristic function scaling exponentially with the number of agents. This would require solving the Vehicle Routing Problem (an NP-Hard problem) an exponential number of times. We therefore propose to model this problem as a coalitional bargaining game where \--- crucially \--- agents are \emph{not} given access to the characteristic function. Instead, we \emph{implicitly} reason about the characteristic function, and thus eliminate the need to evaluate the VRP an exponential number of times \--- we only need to evaluate it once. Our contribution is that our decentralised approach is both scalable and considers the self-interested nature of companies. The agents learn using a modified Independent Proximal Policy Optimisation. Our RL agents outperform a strong heuristic bot. The agents correctly identify the optimal coalitions 79\% of the time with an average optimality gap of 4.2\% and reduction in run-time of 62\%.   
\end{abstract}

\section{Introduction}
\label{sec: intro}

Heavy Goods Vehicles (HGVs) in the UK contributed 4.3\% of the UK's \emph{total} greenhouse gas emissions in 2019 \citep{uk_department_for_business_energy__industrial_strategy_final_2021}. Furthermore, HGVs are utilised inefficiently at 61\% of their total weight capacity. Moreover, 30\% of the distance travelled is empty \citep[RFS0125]{uk_department_for_transport_road_2020}.

\begin{figure}
  \captionsetup[subfigure]{justification=centering}
  \setlength\belowcaptionskip{-12ex}
  \centering
  \subfloat[Pre-collaboration \newline Total Cost = 3.35]{\includegraphics[width=0.3\linewidth]{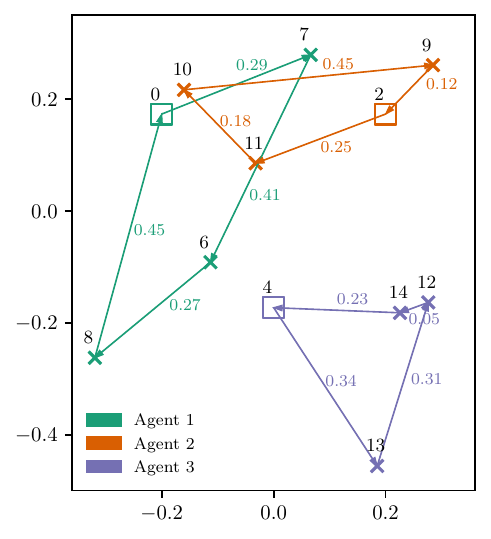}\label{fig: f1a}}
  \subfloat[Post-collaboration (with grand coalition, $\{1, 2, 3 \}$) \newline Total Cost = 2.47]{\includegraphics[width=0.295\linewidth]{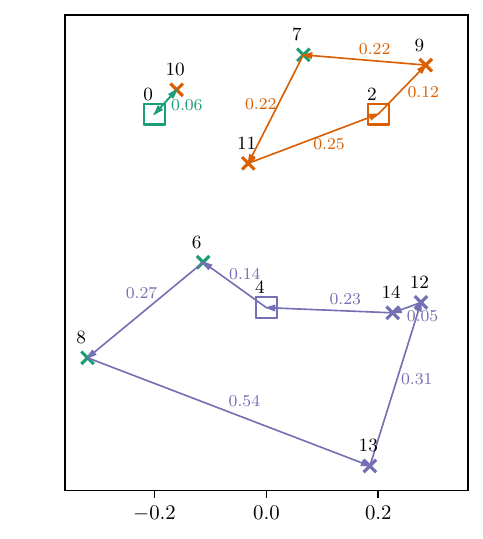}\label{fig: f1b}}
  \hspace{1em}
  \caption{Three agents (denoted by colours) before and after collaboration. Squares denote depots. Crosses denote customer locations. Node indices (arbitrary) are denoted in black, with costs given in their respective colours. The \emph{collaboration gain} is defined as the difference in social welfare before and after collaboration. In Figure \autoref{fig: f1b}, Agents 1, 2 and 3 decide to cooperate forming the grand coalition and reduces the system's total cost by 0.88 or 26\%. This results in a \emph{collaboration gain per capita} (assuming agents split the gain equally) of 0.29. For detailed calculations, see \autoref{app_a}.}
  \label{fig: f1}
\end{figure}

Collaborative Vehicle Routing (CVR) has been proposed to improve HGV utilisation. Here, delivery companies, or \emph{carriers}, need to share their delivery information in order to achieve economies of scale. If companies agree to work together, they are said to be in a \emph{coalition}. The case where all companies join the same coalition is called the \emph{grand coalition}. As a result of improved utilisation, total travel costs can be reduced resulting in a \emph{collaboration gain}. The remaining question is then how to allocate this collaboration gain in a reasonable manner so that carriers are incentivised to form coalitions. An example of Collaborative Vehicle Routing is given in \autoref{fig: f1}.

Prior literature suggests that collaborative routing can reduce costs by around 20-30\% with further reductions to greenhouse gas emissions and road congestion \citep{cruijssen_joint_2007, montoya-torres_impact_2016, gansterer_collaborative_2018, gansterer_shared_2020, ferrell_horizontal_2020}. However, real-world adoption remains limited, with only a few companies participating, according to a number of case-studies conducted \citep{cruijssen_joint_2007, ballot_reducing_2010, cruijssen_horizontal_2007-1, guajardo_review_2016}. One reason behind this is  the complexity of fair \emph{gain sharing} amongst a larger number of companies.

Whilst gain sharing has been studied in collaborative routing using cooperative game theory \citep{guajardo_review_2016}, the solution concepts typically assumes that the characteristic function is given. For a set of $n$ agents, $N = \{1, \dots, n\}$, the characteristic function $v: \mathbf{2}^N \to \mathbb{R}$ assigning a \emph{value}, or in our case \emph{collaboration gain}, for every possible coalition that could be formed. Note that there exists $\mathcal{O}(2^n)$ possible coalitions. This is intractable for settings with more than a few agents, because evaluating the collaboration gain of even a single coalition, involves solving a Vehicle Routing Problem (VRP) \citep{dantzig_truck_1959,toth_vehicle_2014} which is NP-Hard. For detailed calculations of the collaboration gain, see \autoref{app_a}.

Our first contribution is modelling the collaborative routing problem as a \emph{coalitional bargaining game} \citep{rubinstein_perfect_1982, okada_noncooperative_1996}. Here, agents attempt to reach agreement on how to divide a ``pie'' (e.g. \$100) between them through multiple rounds of bargaining (see \autoref{sec: background} for a formal definition). A benefit of this approach is that we consider both the routing problem (Who should deliver which requests?) and the gain sharing problem (Who receives how much of the added value?) simultaneously, whereas a key limitation of many previous methods consider these sub-problems in isolation from one another \citep{gansterer_collaborative_2018}. Moreover, our approach is agnostic to the underlying routing problem \--- the complexity of the VRP formulation could be increased by further constraints such as  time-windows, without further modification to the method.

Our second contribution is that agents do not need access to the characteristic function explicitly. Instead, they can reason about the characteristic function through only receiving a high-dimensional graph input of delivery information (for example, latitudes and longitudes), as well as other agents' actions. This eliminates the need to fully evaluate the characteristic function, which involves solving an NP-Hard problem an exponential number of times. Instead, we only need to solve this NP-Hard problem once when deployed in real-world settings, thus allowing our approach to scale.

To evaluate performance, we measure the agents' ability to select the optimal coalitions. The optimal coalition for player $i$ is defined as the coalition that maximises the value player $i$ receives after sharing gains amongst all agents in the coalition. Firstly, we measure the accuracy of agents to propose the optimal coalitions. Next, we also measure the average optimality gap between the value of their proposed coalition and the value of the optimal coalitions. 

The work of \citep{okada_noncooperative_1996} analyses this coalitional bargaining game (but not in a collaborative routing setting), and proves that agents will cooperate by sharing gains equally (i.e. the egalitarian split) in our setting. Due to the early-stage nature of this paper, we first focus on developing strong RL agents that can identify and join the optimal coalitions. This is already challenging as it requires agents to reason about the characteristic function through receiving a high-dimensional input. We simplify the action space by assuming all agents adopt the egalitarian split as expected by \citep{okada_noncooperative_1996}. We leave it to future work where agents can also learn this aspect themselves.

\section{Related Work}
\label{sec: related_work}
Prior collaborative routing literature tackles the partner selection sub-problem (i.e. who should each carrier work with?) by estimating the collaboration gain between different partners (or companies) \citep{palhazi_cuervo_determining_2016, adenso-diaz_analysis_2014}. However, a limitation of this approach is that they do not consider how much each agent should be compensated, nor if agents even agree to join the same coalitions i.e. if the coalitions are stable. Therefore, posing this problem as a \emph{coalitional bargaining} game aims to rectify this limitation.

The majority of the Collaborative Routing literature is concerned with the exchange of \emph{individual} delivery requests amongst the carriers. This can be divided into three types of planning approaches: centralised; decentralised without auctions; and decentralised with auctions \citep{gansterer_collaborative_2018, gansterer_shared_2020}.

Centralised planning approaches desire to simply maximise social welfare. Typically, this goal is achieved by using a form of mixed integer linear programming or (meta)heuristics \citep{cruijssen_joint_2007,lin_cooperative_2008, montoya-torres_impact_2016}. This can be viewed as a common-payoff setting where all agents receive the same reward.  However, assuming a common-payoff setting in practice is restrictive as companies are \emph{self-interested} \--- they mostly care only about their own profits \citep{cruijssen_horizontal_2007}. Therefore, a more realistic setting is that of decentralised control with self-interested agents.

A key challenge in decentralised settings is managing the explosion in the number of \emph{bundles}. Consider \autoref{fig: f1} where Agent 2 may desire to sell delivery node 10 perhaps to Agent 1. However, if Agent 2 offers both nodes 10 and 11 as a \emph{bundle}, then Agent 2 may be able to command a higher price. Indeed, the number of possible bundles scales $\mathcal{O}(2^m)$ where $m$ is the number of deliveries. To manage this explosion, a heuristic is typically implemented where agents can only submit or request a few bundles (sometimes only one) which would severely impact optimality \citep{bo_dai_mathematical_2009, li_request_2015, xu_truthful_2017}.

A second challenge is to also elicit other agents' preferences over all bundles. One approach is to invoke structure on the problem in the form of \emph{combinatorial auctions} \citep{cramton_combinatorial_2006} which aids optimality \citep{krajewska_horizontal_2008,gansterer_collaborative_2018, gansterer_cost_2019}. However, this additional structure comes at additional computational complexity. Moreover, in auction mechanism design, there are four desirable properties: efficiency; individual rationality; incentive compatibility; and budget balance. \citep{gansterer_cost_2019} shows that the Vickrey-Clarke-Groves Auction and both of their approaches are unable to satisfy all four properties simultaneously and there exists a trade-off instead.

Coalition formation has also been extensively studied in cooperative game theory \citep{chalkiadakis_computational_2011, wooldridge_introduction_2009, shoham_multiagent_2008}. However, much of the existing literature assumes that the characteristic function is given. Alternatively, they aim to find more succinct representations of the characteristic function, typically at a cost of increased computational complexity when computing solution concepts \citep{wooldridge_introduction_2009}. Examples include Induced Subgraph Games and Marginal Contribution Nets \citep{deng_complexity_1994, ieong_marginal_2005}; however, even these representation schemes require evaluating the value of multiple coalitions (and thus solving multiple VRPs). We argue that many real-world scenarios consist of the characteristic function being a function of the agents' assets or capabilities \--- in our case, a function of the deliveries an agent possesses. We therefore ask:\emph{ ``Can agents form optimal coalitions from the delivery information instead of having access to the characteristic function?''}. Therefore, our paper can be viewed as using an alternative, succinct representation scheme and approximating a rational outcome by using a function approximator.

The most similar work to ours is that of \citep{bachrach_negotiating_2020,chang_multi-issue_2020}. In \citet{bachrach_negotiating_2020}, they apply Multi-Agent Reinforcement Learning (MARL) to a spatial and non-spatial Weighted Voting Game, where agents are given access to the characteristic function (in this case, all agents' weights and the quota, $q$). In \citet{chang_multi-issue_2020}, they apply MARL to the 2-player multi-issue bargaining game. Here, both agents' weights over the multiple issues are held constant and thus agents do not require the weights as explicit input to the agents. In our work, we consider the three agents case (which can be extended to the $n$-player setting), where the joint policy space is richer, potentially allowing for collusion. However, for simplicity, we focus only on the single-issue setting.

Finally, significant progress has been made in applying single-agent RL to Travelling Salesman and Vehicle Routing Problems \citep{vinyals_pointer_2017,nazari_reinforcement_2018,kool_attention_2019, joshi_efficient_2019, lu_learning-based_2020, wang_rewriting_2021}. We build upon prior work by adopting the neural network design of \citep{kool_attention_2019} trained via Independent Proximal Policy Optimisation (I-PPO) \citep{schulman_proximal_2017, de_witt_is_2020} with a few modifications due to the high computational cost of stepping our environment. Further details on agent design are given in \autoref{sec: agent_design}.

\clearpage

\section{Background}
\label{sec: background}

\paragraph{Coalitional Games}
We consider the $n$-player coalitional game (or cooperative game) with a set of agents $N = \{1, \dots, n\}$. A \emph{coalition} is defined as a subset of N, i.e. $S \subseteq N$. The set of all coalitions is denoted $\Sigma$. The \emph{grand coalition} is where the coalition consists of all agents in N, i.e. $S = N$. A \emph{singleton coalition} is where the coalition consists of only one agent, i.e. $|S| = 1$.

A (transferable utility) coalitional game is a pair $G = \langle N, v \rangle$. The \emph{characteristic function} $v: \mathbf{2}^N \to \mathbb{R}$ represents the \emph{value} (or collaboration gain in our setting) that a given coalition $S$ receives. Like \citep{okada_noncooperative_1996}, we assume that the characteristic function is \emph{0-normalised}, \emph{essential} and \emph{super-additive}. The characteristic function is \emph{0-normalised} if the value of all singleton coalitions is 0, i.e. $v(\{i\}) = 0, \forall i \in N$. It is \emph{essential} if the value of the grand coalition is strictly positive, $v(N) > 0$. It is \emph{super-additive} if $v(S \cup T) \geq v(S) + v(T)$ for all coalition pairs $(S, T) \in \Sigma$ where $S \cap T = \varnothing$.

The payoff vector $\mathbf{x}^S = \langle x_1, \dots, x_k \rangle, \ \sum_i{x_i} = 1, \ x_i \in [0, 1]$ denotes how the value achieved by a coalition $S$ is distributed amongst its $k$ members.  The payoff vector is \emph{feasible} if $\sum_{i \in S} x_i^S \leq v(S)$. The set of all feasible payoff vectors for a given coalition S is $X^S$, and $X_+^S$ when all the elements of $X^S$ is non-negative.

\paragraph{Coalitional Bargaining} \citep{okada_noncooperative_1996} presents the $n$-player, random proposers, alternating offers coalitional bargaining game which we adopt. For simplicity, we focus on the 3-player setting. At every timestep  $t = 1, 2, \dots$ an agent from N is selected uniformly at random to be the \emph{proposer}.  The proposer, player $i$, has two actions. Firstly, to propose a coalition $S$ with $i \in S \subseteq N$ and $v(S) > 0$. Due to the characteristic function being 0-normalised this implies $|S| \geq 2$. Secondly, the proposer proposes a payoff vector $\mathbf{x}^S \in X_+^S$. We currently assume the payoff vector to be an egalitarian split \--- all agents within coalition $S$ distribute the value equally. Next, the remaining agents called the \emph{responders} decide to either accept or reject the proposal. If all agents in the proposed coalition $S$ accepts, then the episode terminates with agent $i$ receiving a reward of $v(S) \cdot x_i$ if $i$ is in the coalition $S$, and 0 otherwise (as the game is 0-normalised). If any responder agents in $S$ rejects the proposal, then the next round of bargaining begins with a new proposer selected uniformly at random and the timestep incremented by 1. Since a discount factor of $\gamma$ is applied which decreases the value received, this encourages agents to reach agreement within the first timestep as shown in \citep{okada_noncooperative_1996}. The discount factor in this setting is analogous to the \emph{patience} of an agent, or the urgency of the delivery decision. The episode continues until either agreement is reached, or the finite time horizon is reached. If the episode is terminated due to reaching the finite time horizon, all agents receive a reward of 0. 

We note that whilst \citeauthor{okada_noncooperative_1996}'s results assumes the infinite horizon case, like in \citeauthor{bachrach_negotiating_2020}, we assume a finite time horizon as \citeauthor{okada_noncooperative_1996} shows that agreement should be reached without delay, i.e. at time $t = 1$. This is primarily due to the use of the discount factor $\gamma$ which decreases the overall collaboration gain that could be achieved as the number of rounds increases. Therefore, agents are incentivised to reach an agreement as soon as possible.

\paragraph{Multi-Agent Reinforcement Learning} We model this problem as an $n$-player \emph{Stochastic Game} i.e. assuming perfect information over states, but not other agents' preferences. This can be defined as a tuple $\langle N, S, A, \mathcal{T}, \mathcal{R} \rangle$ where:

\begin{itemize}
  \item $N$ denotes the set of $n$ agents
  \item $S$ denotes the set of states
  \item $A = A_i \times \dots \times A_n$ denotes the set of joint actions, where $A_i$ is player $i$'s set of actions.
  \item $\mathcal{T}: S \times A \to S$ denotes the transition probabilities
  \item $\mathcal{R}: S \times A \times S \to \mathbb{R}$ denotes the reward function
\end{itemize}

For every timestep $t$, an agent $i$ observes the global state $s$ and outputs an action $a_{i, t}$ sampled from its policy $\pi_{\theta_i}(a_{i, t} \mid s_t)$ which is parameterised by $\theta_i$. This results in a reward $r_{i, t} = \mathcal{R}(s_{t}, \mathbf{a}, s_{t + 1})$. The return $G_i$ is discounted by a factor $\gamma$, given by $G_i = \sum_{t = 1}^{T} \gamma^{t-1} r_{i, t}$. Agent $i$'s objective is to maximise its expected discounted sum of rewards $\mathbb{E}[\sum_{t=1}^T \gamma^{t-1}r_{i, t}]$.

\section{Agent Design}
\label{sec: agent_design}

\paragraph{Input Space} The agents receive a variety of input from the state: 

A \emph{location} can refer to either a depot location or a customer location. A location is defined as a 4-tuple $\langle x, y, \texttt{owner}, \texttt{is\_depot} \rangle$. Indeed, $x$ and $y$ refers to the $x, y$ coordinates of the location. The \texttt{owner} is an integer denoting to which agent the location belongs to. \texttt{is\_depot} takes the value 1 if the location is a depot, and 0 if the location is a customer.

Note that we consider perfect information, that is, each agent sees information over all depots' and customers' locations (including those that are owned by other agents). However, agents do not have access to other agents' preferences.

The agents also receive the current timestep $t$, or bargaining round, as input. In addition, the actions taken by other agents are received as well, denoted by \texttt{actions\_taken}. This has a constant shape of $\langle \texttt{max\_time\_horizon}, \ 2 \times \texttt{n\_agents} \rangle$, where \texttt{max\_time\_horizon} is the maximum number of timesteps before the episode is forcefully terminated, and \texttt{n\_agents} the number of agents. The \texttt{actions\_taken} matrix is initialised all as $-1$. The first three columns are populated with the proposed pay-off vectors at each round of bargaining. The last three columns are populated with the responses of each agent at each round of bargaining.

\paragraph{Actor Feature Extractor Design} The feature extractor is based on a modified version of \citep{kool_attention_2019} as their model achieves high performance across a range of routing problems. Similar to \citep{kool_attention_2019}, we use an embedding dimension of 128.

The \texttt{location} information is fed through three encoder layers of a Transformer \citep{vaswani_attention_2017} to form our node embeddings. However, \citep{kool_attention_2019} finds that replacing layer normalisation with batch normalisation improves performance and thus we adopt batch normalisation as well. To obtain our graph embedding, we take a mean over the node embeddings. We additionally embed the timestep $t$ into a 128-dimensional vector to provide context, which we sum with the node embeddings. We then take a single-head attention over the node embeddings to form a resultant embedding. The resultant embedding is then used to decide which actions to take.

\paragraph{Action Space} The agents have three action heads: \emph{coalitions}, \emph{proposals} and \emph{response}.

\texttt{coalitions} takes the form $\{0, 1\}^{|N|}$ where $|N|$ is the total number of agents, in this case, 3. This action denotes whether the respective agent is part of the coalition $S$. Note that this game assumes that player $i$ is in the coalition $S$, i.e. $S_i = 1$. The \texttt{coalitions} action head takes the resultant embedding followed by four dense layers with 256 hidden neurons and a ReLU activation function. The output is then passed through $n$ independent Bernoulli distributions to determine the probability that a given agent exists within the coalition $S$. 

\texttt{proposals} is a vector $\mathbf{x} \in \mathbb{R}_+^{|N|}$ where $\sum_i{x_i} = 1, \ x_i \in [0, 1]$. This vector denotes how much of the collaboration gain is assigned to each respective agent (as a percentage). Note that this action head is not strictly necessary in this current paper as we assume an egalitarian split; however, we would require this in future work if agents are to learn the gain sharing mechanism themselves. Furthermore, note that this is a continuous action space, as opposed to the other actions which are discrete. To calculate \texttt{proposals}, the resultant embedding is passed through a single dense layer with 64 hidden neurons and a $\tanh$ activation function. From this, a mean and standard deviation is output and fed into a multi-variate Guassian. This Guassian is then sampled from to form the logits of the pay-off vector $\mathbf{x}$. This pay-off vector is then masked by the \emph{coalitions} vector, i.e. if a player $i$ is not in the coalition $S$, it will also receive 0 in the pay-off vector. An egalitarian split is forced by masking the logits to be 1 if the respective agent is in the proposed coalition $S$, and an arbitarily large negative number otherwise. Finally, a softmax is taken over the logits to ensure the constraint $\sum_i{x_i} = 1$.

\texttt{responses} is a binary action that denotes whether an agent accepts or rejects a given proposal. It takes the resultant embedding followed by four dense layers with 256 hidden neurons and a ReLU activation function. The output is then fed through a Bernoulli distribution.

\paragraph{Loss Function} The agents are trained through independent learning \citep{tan_multi-agent_1993} in the form of Independent PPO \citep{schulman_proximal_2017, de_witt_is_2020}. However, modifications had to be made. Firstly, we use a baseline instead of a critic (i.e. no bootstrapping in value function estimates) as the design of a good critic for vehicle routing problems is ``non-trivial'' \citep{kool_attention_2019}. To perform the greedy roll-out proposed by \citep{kool_attention_2019} is expensive due to the computational cost of stepping our environment. Thus, we use the baseline in \citep{nazari_reinforcement_2018} which estimates the value only of the initial state, $S_0$ (see below paragraph on baseline design \ref{para: baseline}); this has an appealing interpretation as the \emph{difficulty} of a given problem instance. For simplicity, we do not use shared parameters for the actor and baseline to avoid tuning an additional hyperparameter, albeit potentially at the cost of sample efficiency. Note that we also normalise the advantage function due to the small magnitude in rewards.

The loss function is the clipped objective function of \citep{schulman_proximal_2017} with entropy regularisation:

\[
\mathcal{L}_t^{\text{CLIP}}(\theta) = \hat{\mathbb{E}_t} [\min (r_t(\theta)\hat{A}_t, \, \text{clip}(r_t(\theta), 1 - \epsilon, 1 + \epsilon) \, \hat{A}_t) + \beta S[\pi_\theta](s_t)]
\]

where $\beta$ is the entropy regularisation coefficient and $S$ the entropy bonus. Strong entropy regularisation was employed. The entropy regularisation coefficient $\beta$ was linearly annealed from 0.75 to 0 over 10,000 epochs, followed by $\max(\beta, 0.2)$. Such high entropy regularisation was essential in this setting, as a strong locally optimal policy would be to always propose the grand coalition and always accept every proposal. Without entropy regularisation, the agents would converge on this local optimum too quickly and agents would not learn to extract useful features over the states.

Due to computational limitations, a hyperparameter search was not performed. We suspect that small improvements to sample efficiency could be made by finding more optimal hyperparameter settings and schedules.

\paragraph{(Reward) Baseline Design}
\label{para: baseline}
A useful baseline helps reduce the variance in policy-gradient methods. The baseline we use estimates the value of the initial state $V_{\pi_\theta} (S_0)$. This has an appealing interpretation as being approximately the difficulty of a given problem instance \citep{nazari_reinforcement_2018}. The baseline adopts the same neural network architecture as the actor to arrive at the resultant embedding. However, we then feed this embedding through six dense layers of 256 hidden neurons using a ReLU activation function. The baseline is trained using a clipped mean squared error loss as in \citep{schulman_proximal_2017}.

\section{Experiments}
\label{sec: exp}

\begin{wrapfigure}{r}{0.25\textwidth}
  \vspace{-5mm}
  \setlength\belowcaptionskip{-2ex}
  \centering
  \includegraphics[width=\linewidth]{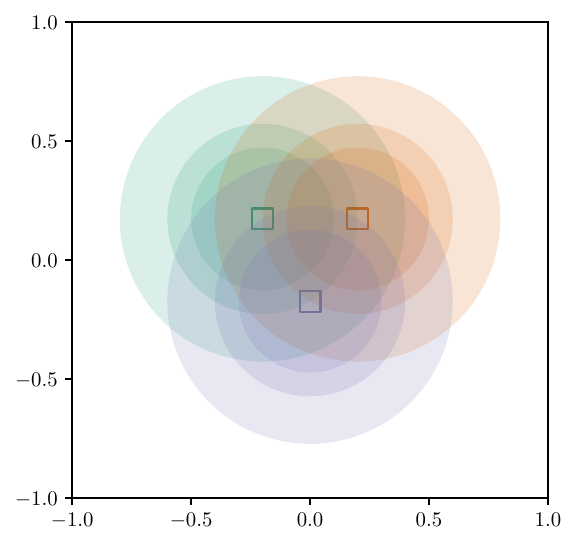}  
  \caption{A plot of the distribution of depot and customer locations. Depots are fixed denoted by squares. Customers may be uniformly at random located within any of agent's respective circles.}
  \label{fig: f2}
\end{wrapfigure}

\paragraph{Problem Setting} We base our problem setting on a modified version of \citep{gansterer_centralized_2018}. We consider an environment with three companies, each represented by an agent. Each agent has one depot and three customers that it must deliver to. The depot $(x, y)$ locations for each agent are held fixed at $\{(-0.2, 0.173), (0.2, 0.173), (0, -0.173)\}$ respectively. The customer $(x, y)$ locations are generated uniformly at random within a certain radius. The radius for each instance is selected uniformly at random from the set $\{0.3, 0.4, 0.6\}$. Varying the radius has the effect of varying the degree of overlap between agents \--- and thus opportunity for collaboration. This can be seen in \autoref{fig: f2}. Future work will consider an increased number of agents and real-world delivery data.

To calculate the pre-collaboration and post-collaboration gains, the shortest paths are calculated exactly using Gurobi \citep{gurobi_optimization_llc_gurobi_2021}. The pre-collaboration shortest paths can be calculated by solving three (un)Capacitated Vehicle Routing Problems (one for each agent). The post-collaboration shortest paths are calculated by solving a single Multi-Depot Vehicle Routing Problem. Problem formulations for the Capacitated VRP and Multi-Depot VRP can be found in \autoref{app: b} and \autoref{app: c} respectively. Capacity is effectively removed by setting the capacity of each vehicle to an arbitrarily large number and the weight of each delivery to 1.

\paragraph{Experimental Design} We perform 5 independent runs with different random seeds to train our agents. Agents are trained for 10,000 epochs and evaluated every 100 epochs. Every training instance is randomly generated on the fly \--- thus agents never see the same instance twice, both during training and evaluation. We train using a batch size of 256 and evaluate with a batch size of 512. All agents use a discount factor $\gamma$ of 0.99. The maximum number of bargaining rounds $T$ is set to 10. The learning rate was held constant at \num{3e-4} and we use Adam \citep{kingma_adam_2017}. All code is run on a single desktop with an Intel\textregistered\ Core\texttrademark\ i7-9700K processor with an NVIDIA\textregistered\ GeForce\textregistered\ RTX 2080Ti GPU. We use TensorFlow 2.4.1 \citep{martin_abadi_tensorflow_2015} to implement our agents.

\paragraph{Heuristic Bot} To compare our algorithm with a simple baseline algorithm, we hand-craft a bot that always forms the grand coalition and always accepts every proposal. This bot has two interpretations. Firstly, due to the slight class imbalance, this is simply a bot that always selects the majority class for each agent in the proposed coalition. An alternative interpretation is that this bot can be viewed as an algorithm with a common-payoff. This has the effect of maximising the collaboration gain (as this game is super-additive), but may not always provide stable coalitions (it may be rational for some agents to defect from a given coalition). This can be most easily illustrated in settings where a \emph{dummy player} exists. A dummy player is a player who adds no value to a given coalition. An obvious example would be a player $i$ who is situated far away from the other two players $j$ and $k$ in the collaborative routing setting.

\subsection{Evaluation Metrics}
\paragraph{Accuracy} A simple evaluation metric is to measure how often the agents propose the correct coalition. For player $i$, the correct coalition $S_i^*$ is defined to be the coalition $S$ which would maximise player $i$'s reward. This involves brute forcing the characteristic function to evaluate the value of each possible coalition which is only possible since we consider 3 agents. The reward $R$ is the collaboration gain from agreeing to coalition $S$, $v(S)$, multiplied by the $i$th element of the pay-off vector, $x_i$.

\paragraph{Optimality Gap} We denote the absolute and relative optimality gap of player $i$ by $\phi_i$ and $\eta_i$ respectively. The absolute optimality gap $\phi_i$ for player $i$ is defined as $\phi_i = v(S_i^*) - v(S)$, where $S_i^*$ is the correct coalition, $S_i$ is player $i$'s proposed coalition, and $v(\cdot)$ is the characteristic function (i.e. the collaboration gain of a given coalition). The relative optimality gap $\eta_i$, is calculated as $\eta_i = \frac{v(S_i^*) - v(S_i)}{v(S_i^*)}$.

Since the data is randomly generated, there could be scenarios where there is no value in collaborating, i.e. even the value of the grand coalition is 0, $v(N) = 0$. Note that we exclude these scenarios when calculating the above evaluation metrics; however, this only occurs 1.9\% of the time when brute-forcing 51,200 instances.

\subsection{Results}
\begin{figure}[h!]
  \centering
  \subfloat[Average Accuracy (\%)]{\includegraphics[width=0.425\linewidth]{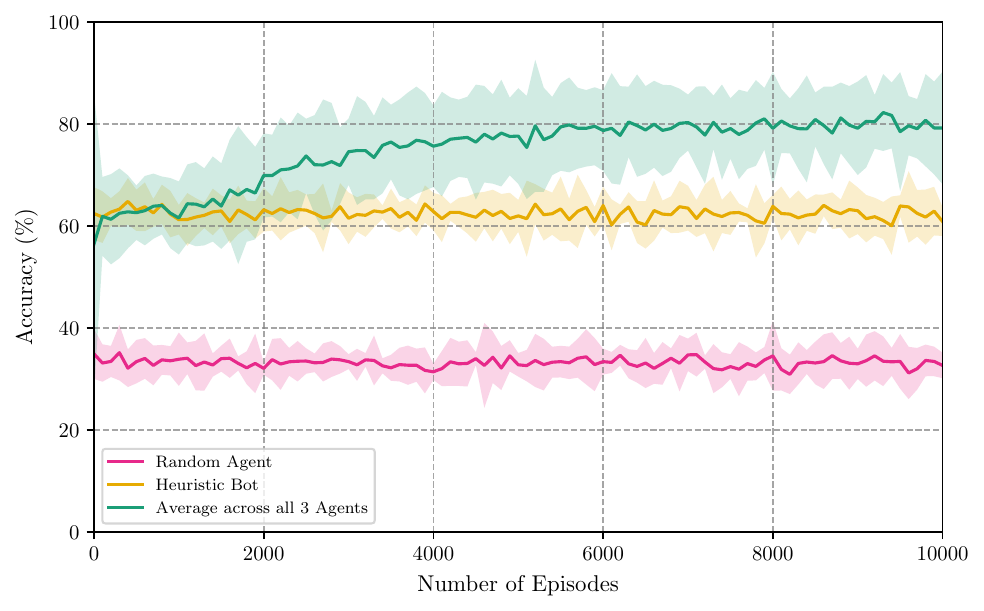} \label{fig: f3a}} 
  \subfloat[Average Optimality Gap (\%)]{\includegraphics[width=0.425\linewidth]{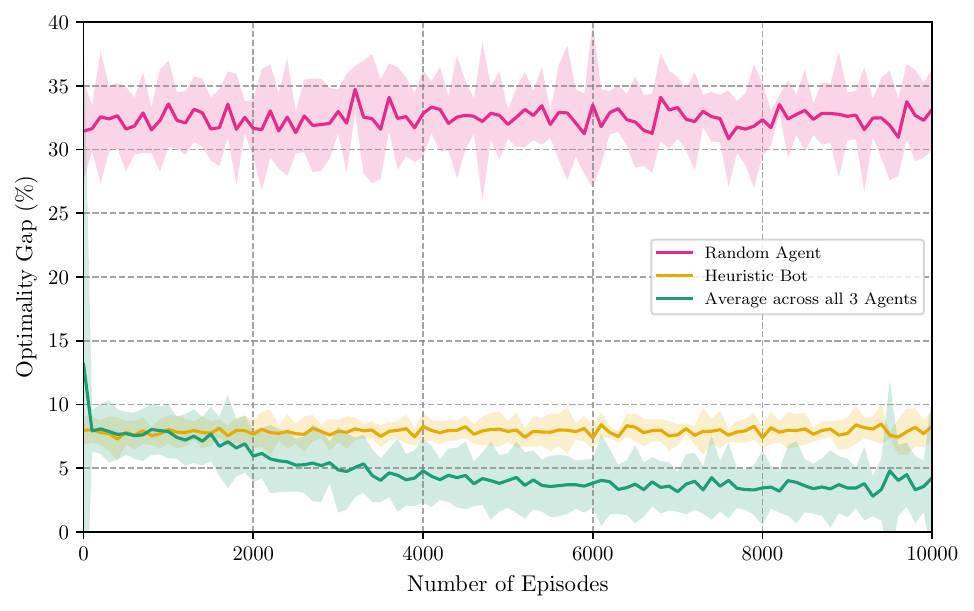} \label{fig: f3b}}
  \caption{Learning curve taking an (a) average accuracy (b) average optimality gap across all 3 agents for readability. For individual learning curves of each agent, see \autoref{app: d} and \autoref{app: e}. Solid lines denote mean accuracy across all 15 runs (five for each of the three agents). Shaded regions denote $\pm$ two standard deviations. After training for 10,000 epochs, our RL agents reach an average accuracy of 79\%, outperforming the heuristic bot.}
  \label{fig: f3}
\end{figure}

We perform 5 independent runs comparing our RL bot to the heuristic bot, as well as a random agent which simply proposes coalitions uniformly at random. We evaluate the performance of our agents both in terms of accuracy, optimality gap and run-time. This can be seen in \autoref{fig: f3}.

\clearpage

From Figures \ref{fig: f3a} and \ref{fig: f3b}, we conclude that our agents have learnt to outperform the heuristic bot reaching an average accuracy of 79\% and average optimality gap of 0.01 (or 4.2\%). For individual learning curves of absolute optimality gap, see \autoref{app: f}. Furthermore, our RL agents are able to reach agreement in 512 parallel instances within an average of 9.2s (or 0.018s per instance). We note that to brute force the characteristic function for 512 instances takes 24.3s (or 0.047s per instance). Thus, our RL agents achieve a 62\% reduction in computational time. Whilst 0.047s per instance may seem reasonable even when applying brute force, we stress that this is due to the simplistic VRP setting we consider \--- this will not scale with the number of agents nor problem complexity via additional constraints such as time-windows. Moreover, our agents reach agreement in a distributed and self-interested manner, which overcomes the limitations of central orchestration methods mentioned in \autoref{sec: related_work}. We hypothesise that a key remaining challenge is due to the low magnitude in absolute optimality gap of 0.01. To overcome this, future work could investigate this challenge from either an algorithmic improvement point of view, or simply use a more realistic revenue and cost structure for the underlying routing problem (where the magnitude is much greater). In addition, there exists symmetries in our delivery information. Future work could investigate learning more robust representations over the input, such that the actions taken are invariant to linear transformations, such as rotation and reflection.

It is also difficult to parallelise this workflow efficiently when calculating the collaboration gains as it is all performed on CPU. It should be possible to sacrifice the mathematical guarantee of optimality with approximate solution methods, such as (meta)heuristics, or indeed \--- reinforcement learning. Using RL in the environment would allow us to easily shift the computation to the GPU and should enable a larger number of instances to be solved in parallel. However, it would be important to ensure the robustness of these RL models first and thus we leave it for future work.

\section{Conclusion}
In this paper, we propose to tackle the challenging real-world problem of Collaborative Vehicle Routing through a coalitional bargaining lens with reinforcement learning. The main challenge of our setting is the inability of extant methods to fully evaluate the characteristic function due to high computational complexity. The RL agents designed in this work are able to correctly reason over a high-dimensional graph input to \emph{implicitly} reason about the characteristic function instead. This eliminates the need to evaluate the Vehicle Routing Problem (VRP) an exponential number of times and increase its practicability as we only need to perform this once. Another important point is that collaboration is not centrally orchestrated but facilitated using distributed decision making. This marks an important step towards real-world adoption which would aid transportation planners to consider more possible collaboration scenarios.

The current focus of this work is to obtain strong RL agents that can identify and join the optimal coalitions. Whilst we have achieved this, a key current assumption in our work is the egalitarian split (where agents split the collaboration gain equally). Instead, in future work agents should learn this themselves. Furthermore, future work should investigate more realistic routing scenarios. In particular, capacity is currently represented as a scalar value, whereas real-world settings may require considering the 3-dimensional volume of packages as well. This would lead to a ``Three-dimensional Loading Capacitated Vehicle Routing Problem'' (3L-CVRP).

\section{Broader Impact}
\label{sec: broader}
We believe that tackling collaborative routing through coalitional bargaining and MARL presents a promising research direction which allows reduction in cost, greenhouse gas emissions and road congestion. More broadly, this could be applied to \emph{collaborative logistics} where, for example, companies cooperate by sharing warehouse space and where again gain sharing is essential.

However, there still remains a research gap before this approach can be deployed in the real-world. Firstly, can this system be gamed? Can companies (un)intentionally mis-report the deliveries they possess in the form of \emph{phantom} or \emph{decoy} delivery tasks? More generally, how can we verify companies' information in a sensitive manner such that all parties trust the system? Whilst these are interesting and challenging research questions, we believe that multi-agent reinforcement learning could be a useful tool (in addition to other approaches, such as game theory) to help address these questions.

\begin{ack}
This work was supported by the UK Engineering and Physical Sciences Research Council (EPSRC) grant on ‘‘Intelligent Systems for Supply Chain Automation’’ under Grant Number 2275316, as well as by the UK EPSRC Connected Everything Network Plus under Grant EP/S036113/1.

We thank Eugene Vinitsky for his mentorship which improved the quality of this paper and the Cooperative AI Organising Committee for facilitating the mentorship scheme. We also thank the Manufacturing Analytics Group for their insightful discussions regarding early drafts of this paper. 
\end{ack}

\bibliography{references}
\clearpage

\section*{Checklist}

\begin{enumerate}

\item For all authors...
\begin{enumerate}
  \item Do the main claims made in the abstract and introduction accurately reflect the paper's contributions and scope?
    \answerYes{}
  \item Did you describe the limitations of your work?
    \answerYes{See Abstract and \autoref{sec: intro}}
  \item Did you discuss any potential negative societal impacts of your work?
    \answerYes{See \autoref{sec: broader}}
  \item Have you read the ethics review guidelines and ensured that your paper conforms to them?
    \answerYes{}
\end{enumerate}

\item If you are including theoretical results...
\begin{enumerate}
  \item Did you state the full set of assumptions of all theoretical results?
    \answerNA{}
	\item Did you include complete proofs of all theoretical results?
    \answerNA{}
\end{enumerate}

\item If you ran experiments...
\begin{enumerate}
  \item Did you include the code, data, and instructions needed to reproduce the main experimental results (either in the supplemental material or as a URL)?
    \answerNo{We will publicly release code after the extension of this work to a journal paper.}
  \item Did you specify all the training details (e.g., data splits, hyperparameters, how they were chosen)?
    \answerYes{See \autoref{sec: exp}}
	\item Did you report error bars (e.g., with respect to the random seed after running experiments multiple times)?
    \answerYes{See \autoref{fig: f3}, \autoref{fig: d} and \autoref{fig: e}}
	\item Did you include the total amount of compute and the type of resources used (e.g., type of GPUs, internal cluster, or cloud provider)?
    \answerYes{All code is run on a single desktop with an Intel\textregistered\ Core\texttrademark\ i7-9700K processor with an NVIDIA\textregistered\ GeForce\textregistered\ RTX 2080Ti GPU. A single independent run takes 31.5 hours to train, but we note that after training, it only takes 9 seconds for all agents to reach agreement.}
\end{enumerate}

\item If you are using existing assets (e.g., code, data, models) or curating/releasing new assets...
\begin{enumerate}
  \item If your work uses existing assets, did you cite the creators?
    \answerYes{We cite the use of the neural network design of \citep{kool_attention_2019} and the benchmark design of \citep{gansterer_centralized_2018}. We also thank TensorFlow \citep{martin_abadi_tensorflow_2015} and Gurobi \citep{gurobi_optimization_llc_gurobi_2021}.}
  \item Did you mention the license of the assets?
    \answerNA{}
  \item Did you include any new assets either in the supplemental material or as a URL?
    \answerNo{}
  \item Did you discuss whether and how consent was obtained from people whose data you're using/curating?
    \answerNA{}
  \item Did you discuss whether the data you are using/curating contains personally identifiable information or offensive content?
    \answerNA{}
\end{enumerate}

\item If you used crowdsourcing or conducted research with human subjects...
\begin{enumerate}
  \item Did you include the full text of instructions given to participants and screenshots, if applicable?
    \answerNA{We do not work with human subjects at this early stage of work.}
  \item Did you describe any potential participant risks, with links to Institutional Review Board (IRB) approvals, if applicable?
    \answerNA{}
  \item Did you include the estimated hourly wage paid to participants and the total amount spent on participant compensation?
    \answerNA{}
\end{enumerate}

\end{enumerate}


\appendix
\clearpage
\section{Appendix A - Detailed Calculation of an Optimal Coalition, Collaboration Gain and the Characteristic Function}
\label{app_a}

\begin{figure}[h]
  \captionsetup[subfigure]{justification=centering}
  \centering
  \subfloat[Pre-collaboration \newline Total Cost = 3.35]{\includegraphics[width=0.3\linewidth]{plots/pre_collab_tours.pdf}\label{fig: aa}}
  \hspace{1em}
  \subfloat[Post-collaboration (with grand coalition, $\{1, 2, 3 \}$)  \newline Total Cost = 2.47]{\includegraphics[width=0.295\linewidth]{plots/post_111_tours_no_axis_labels.pdf}\label{fig: ab}}
  \hspace{1em}
  \subfloat[Post-collaboration (with a coalition $\{1, 2\}$ i.e. Agent 3 is excluded from the coalition)  \newline Total Cost = 2.59]{\includegraphics[width=0.295\linewidth]{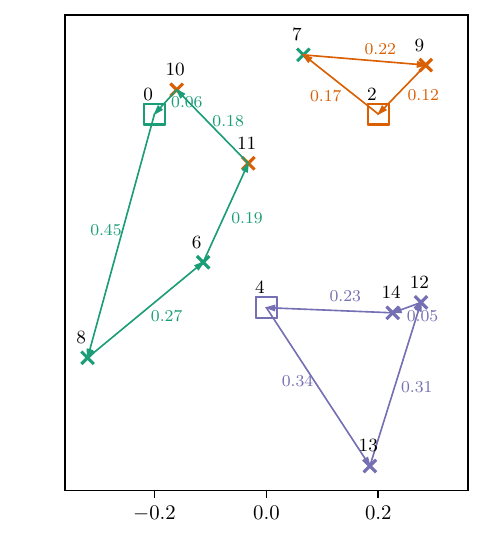}\label{fig: ac}}
  \caption{Three agents, Agents 1, 2 and 3 are denoted by the colours green, orange and purple respectively. Squares denote depots. Crosses denote customer locations. Node indices (arbitrary) are denoted in black, with costs given in their respective colors. The \emph{collaboration gain} is defined as the difference in social welfare before and after collaboration. Figures \autoref{fig: ab} and \autoref{fig: ac} refer to two possible post-collaboration scenarios with collaboration gains per capita of 0.29 and 0.38 respectively. Thus, it would be rational for the coalition $\{1, 2\}$ to form instead of the grand coalition $\{1, 2, 3\}$.}
  \label{fig: a}
\end{figure}

We denote the set of $n$ agents as $N = \{1, \dots, n\}$. A coalition is a subset of $N$, i.e. $S \subseteq N$. The grand coalition is where all agents are in the coalition, i.e. $S = N$. 

\textbf{Pre-collaboration Profit and Social Welfare}: The \emph{pre-collaboration profit} of Agent 1 is calculated as follows: the \emph{Revenue} is 3 (1 for each delivery); the \emph{Cost} is 1.42 (sum of the edge distances); thus the \emph{Profit} is 1.58 (Revenue subtract Cost). Similarly, the pre-collaboration profit of Agents 2 and 3 is 2 and 2.07. The \emph{pre-collaboration social welfare} is the sum of the pre-collaboration profits, thus $1.58 + 2 + 2.07 = 5.65$.

\textbf{Post-collaboration Profit and Social Welfare}: Similarly, the post-collaboration profit of \mbox{Agent 1} is $1 - (0.06 + 0.06) = 0.88$. For Agents 2 and 3 it is 2.19 and 3.46 respectively. Thus a \emph{post-collaboration social welfare} of $0.88 + 2.19 + 3.46 = 6.53$.

\textbf{Collaboration Gain}: The \emph{collaboration gain} is defined as the difference in social welfare before and after collaboration, in this case $6.53 - 5.65 = 0.88$. The \emph{value per capita} is $\frac{0.88}{3} = 0.29$. Note that if only Agents 1 and 2 form a coalition (and exclude Agent 3), then the collaboration gain is divided by 2 instead \--- thus making it rational to object and form the coalition $\{1, 2\}$ (the value per capita of this coalition is 0.38). Due to our assumption of the egalitarian split, i.e. agents share all gains equally, then the reward the agents receive is equal to the value per capita.

\textbf{Characteristic Function}: The characteristic function, $v: \mathbf{2}^N \to \mathbb{R}$ calculates for every possible coalition the collaboration gain. Importantly, to fully evaluate the characteristic function would require solving a variant of the Vehicle Routing Problem for every possible coalition which scales $\mathcal{O}(2^n)$.

Following the example in \autoref{fig: a}:

\begin{center}
  \renewcommand{\arraystretch}{1.2}
  \begin{tabular}{l l}
    $v(\{1, 2, 3\}) = 0.88$ & Value per Capita = $\frac{0.88}{3} = 0.29$ \\
    $v(\{1, 2\}) = 0.76$ & Value per Capita = $\frac{0.76}{2} = 0.38$  \\
    $v(\{1, 3\}) = 0.24$ & Value per Capita = $\frac{0.24}{2} = 0.12$  \\
    $v(\{2, 3\}) = 0.01$ & Value per Capita = $\frac{0.01}{2} = 0.005$  \\
  \end{tabular}
\end{center}

Whilst the grand coalition $\{1, 2, 3\}$ maximises the collaboration gain and thereby social welfare, it is rational for Agents 1 and 2 to object and form the coalition $\{1, 2\}$ instead, as this maximises their own value per capita (and thus reward). Thus, we say that the coalition $\{1, 2\}$ is optimal.

Therefore, if Agents 1 or 2 is proposing a coalition, we measure the accuracy of them correctly proposing the coalition $\{1, 2\}$. For Agent 3, since it is required to propose a coalition that contains itself, yet the optimal coalition does not contain Agent 3, Agents 1 and 2 should object to any proposal from Agent 3. Thus, we exclude these scenarios when calculating the accuracy for Agent 3.

\section{Appendix B - Capacitated Vehicle Routing Problem}
\label{app: b}

In our paper, the pre-collaboration social welfare can be calculated by first solving three independent Capacitated Vehicle Routing Problems, where we assume an arbitrarily high capacity for each vehicle.

The Capacitated Vehicle Routing Problem (CVRP) was first proposed by \citep{dantzig_truck_1959}. Since then, many different variants and formulations of the Vehicle Routing Problem has been proposed \citep{toth_vehicle_2014}. Here we show the \emph{three-index (vehicle-flow) formulation}.

The CVRP considers the setting where goods are distributed to $n$ customers. The goods are initially located at the \emph{depot}, denoted by nodes (or vertices) $o$ and $d$. Node $o$ refers to the starting point of a route, and node $d$ the end point of a route. The customers are denoted by the set of nodes $N = \{1, 2, \dots, n\}$. Each customer $i \in N$ has a \emph{demand} $q_i \geq 0$. In our setting, we consider $q_i = 1$ for all customers. A \emph{fleet} of $|K|$ vehicles $K = \{1, 2, \dots, |K|\}$ are said to be \emph{homogeneous} if they all have the same capacity $Q > 0$. In our setting, we consider only one vehicle and set its capacity $Q$ to an arbitrarily high number to remove the capacity constraint. A vehicle must start at the depot, and can deliver to a set of customers $S \subseteq N$ before returning to the depot. The \emph{travel cost} $c_{i, j}$ is associated for a vehicle travelling between nodes $i$ and $j$ which we assume to be the Euclidean distance.

This problem can be modelled as a complete directed graph $G = (V, A)$, where the vertex set $V \coloneqq N \cup \{o, d\}$ and the arc set $A \coloneqq (V \setminus \{d\}) \times (V \setminus \{o\})$. We define the \emph{in-arcs} of $S$ as $\delta^-(S) = \{(i, j) \in A: i \notin S, j \in S\}$. The \emph{out-arcs} of $S$ is $\delta^+(S) = \{(i, j) \in A: i \in S, j \notin S\}$.

The binary decision variables $x_{ijk}$ denotes whether a vehicle $k \in K$ travels over the arc $(i, j) \in A$. The binary decision variables $y_{ik}$ denotes whether a vehicle $k \in K$ visits node $i \in V$. $u_{ik}$ denotes the load in vehicle $k$ before visiting node $i$. We define the demand at the depot nodes $o$ and $d$ to be 0, i.e. $q_o = q_d = 0$. This yields:

\begin{mini!}
  {}            {\sum_{k \in K} c^T x_k \label{eq: e1o1}}                     {}               {}
  \addConstraint{\sum_{k \in K} y_{ik}}                      {=1, \qquad \label{eq: e1c1}}            {\forall i \in N}
  \addConstraint{x_k(\delta^+(i)) - x_k(\delta^-(i))}        {=\begin{cases} 1, & i = o, \\ 0, & i \in N, \end{cases} \qquad \label{eq: e1c2}}            {\forall i \in V \setminus \{d\}, k \in K}
  \addConstraint{y_{ik}}{= x_k(\delta^+(i)) \label{eq: e1c3}}{\forall i \in V \setminus \{d\}, k \in K}
  \addConstraint{y_{dk}}{= x_k(\delta^-(d)) \label{eq: e1c4}}{\forall k \in K}
  \addConstraint{u_{ik} - u_{jk} + Qx_{ijk}}{\leq Q - q_j \label{eq: e1c5}}{\forall (i, j) \in A, k \in K}
  \addConstraint{q_i}{\leq u_{ik} \leq Q \label{eq: e1c6}}{\forall i \in V, k \in K}
  \addConstraint{x}{=(x_k) \in \{0, 1\}^{K \times A} \label{eq: e1c7}}                         {}
  \addConstraint{y}{=(y_k) \in \{0, 1\}^{K \times V} \label{eq: e1c8}.}                         {}
\end{mini!}

\begin{itemize}
  \item The objective function (\ref{eq: e1o1}) minimises the Euclidean distance travelled by the vehicle.
  \item Constraint (\ref{eq: e1c1}) ensures the vehicle only visits each customer once.
  \item Constraint (\ref{eq: e1c2}) ensures that the sum of vehicles entering node $d$ and exiting node $d$ is $-1$. This ensures that a vehicle $k$ performs a route starting at $o$ and ending at $d$.
  \item Constraint (\ref{eq: e1c3} and \ref{eq: e1c4}) couples variables $x_{ijk}$ and $y_{ik}$.
  \item Constraint (\ref{eq: e1c5}) is the Miller-Tucker-Zemlin constraint which helps eliminate subtours.
  \item Constraint (\ref{eq: e1c6}) is the capacity constraint.
\end{itemize}

\section{Appendix C - Multi-Depot Vehicle Routing Problem}
\label{app: c}

In our paper, the post-collaboration social welfare can be calculated by solving the Multi-Depot Vehicle Routing Problem (MDVRP) once. The number of depots corresponds to the number of agents within the accepted coalition. Again, we remove capacity constraints by setting the capacity of each vehicle to an arbitrarily large number. However, we add the additional constraint that each vehicle has to visit at least one customer.

The MDVRP is a simple extension of the CVRP formulation provided in \autoref{app: b}. Instead of having the depot simply represented by nodes $o$ and $d$, the depots are extended to belong to a specific vehicle $k$ through nodes $o_k$ and $d_k$. Doing so yields:

\begin{mini!}
  {}            {\sum_{k \in K} c^T x_k \label{eq: e2o1}}                     {}               {}
  \addConstraint{\sum_{k \in K} y_{ik}}                      {=1, \qquad \label{eq: e2c1}}            {\forall i \in V}
  \addConstraint{x_k(\delta^+(i)) - x_k(\delta^-(i))}        {=\begin{cases} 1, & i = o_k, \\ 0, & i \in N, \end{cases} \qquad \label{eq: e2c2}}            {\forall i \in V \setminus \{d_k\}, k \in K}
  \addConstraint{y_{ik}}{= x_k(\delta^+(i)) \label{eq: e2c3}}{\forall i \in V \setminus \{d_k\}, k \in K}
  \addConstraint{y_{d_{k}k}}{= x_k(\delta^-(d_k)) \label{eq: e2c4}}{\forall k \in K}
  \addConstraint{y_{d_{k}k}}{= 1 \label{eq: e2c5}}{\forall k \in K}
  \addConstraint{u_{ik} - u_{jk} + Qx_{ijk}}{\leq Q - q_j \label{eq: e2c6}}{\forall (i, j) \in A, k \in K}
  \addConstraint{q_i}{\leq u_{ik} \leq Q \label{eq: e2c7}}{\forall i \in V, k \in K}
  \addConstraint{x}{=(x_k) \in \{0, 1\}^{K \times A} \label{eq: e2c8}}                         {}
  \addConstraint{y}{=(y_k) \in \{0, 1\}^{K \times V} \label{eq: e2c9}.}                         {}
\end{mini!}

\begin{itemize}
  \item The objective function (\ref{eq: e2o1}) minimises the Euclidean distance travelled by all vehicles.
  \item Constraint (\ref{eq: e2c1}) ensures that each vehicle only visits each customer once.
  \item Constraint (\ref{eq: e2c2}) ensures that the sum of vehicles entering node $d_k$ and exiting node $d_k$ is $-1$. This ensures that a vehicle $k$ performs a route starting at $o_k$ and ending at $d_k$.
  \item Constraint (\ref{eq: e2c3} and \ref{eq: e2c4}) couples variables $x_{ijk}$ and $y_{ik}$.
  \item Constraint \ref{eq: e2c5} ensures that each vehicle performs at least one delivery.
  \item Constraint (\ref{eq: e2c6}) is the Miller-Tucker-Zemlin constraint which helps eliminate subtours.
  \item Constraint (\ref{eq: e2c7}) is the capacity constraint.
\end{itemize}

\clearpage

\section{Appendix D - Individual Learning Curves - Accuracy}
\label{app: d}
\begin{figure}[h]
  \centering
  \subfloat{\includegraphics[width=0.675\linewidth]{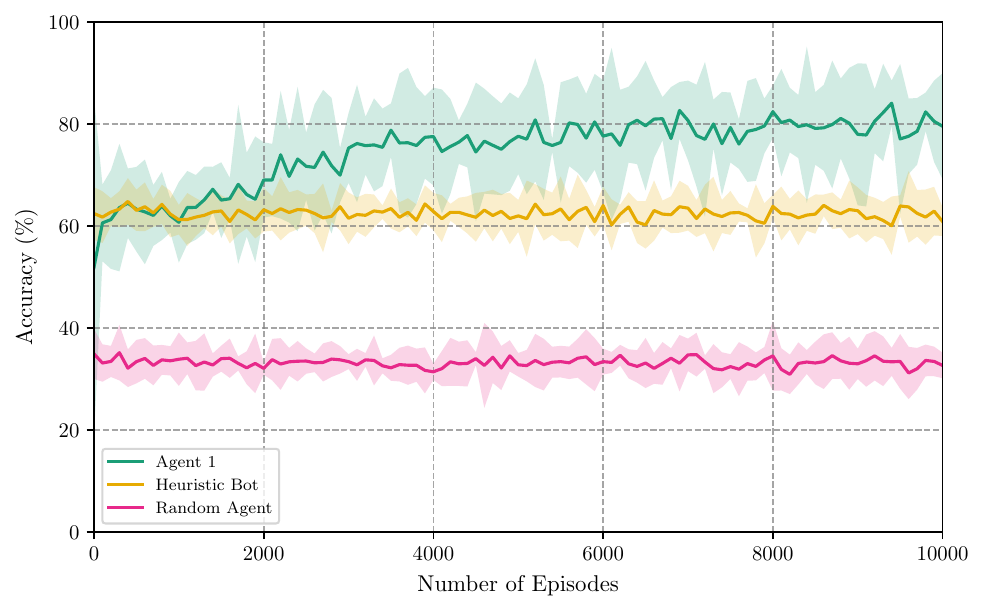}}\label{fig: da}
  \subfloat{\includegraphics[width=0.675\linewidth]{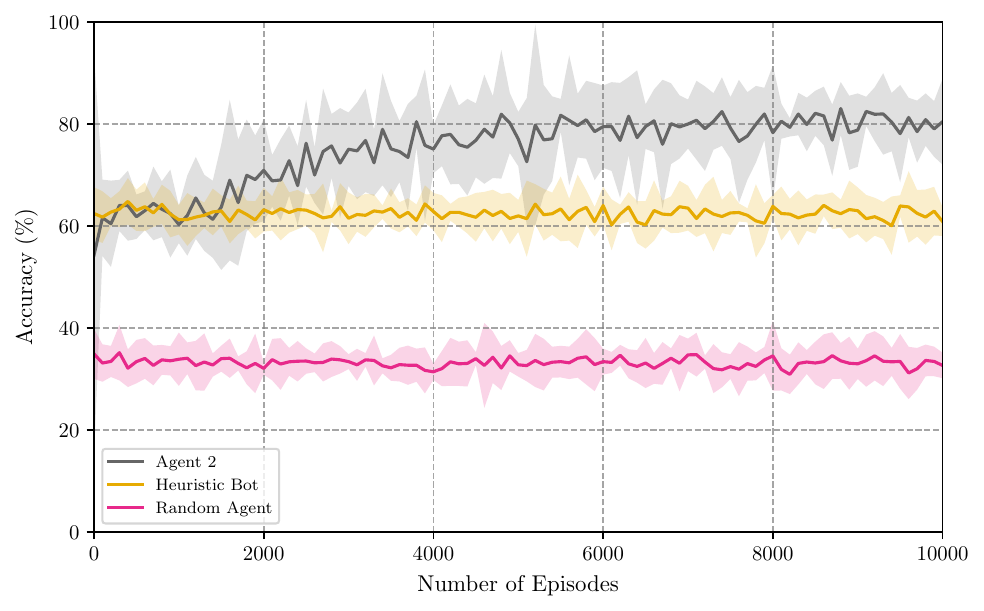}}\label{fig: db}
  \subfloat{\includegraphics[width=0.675\linewidth]{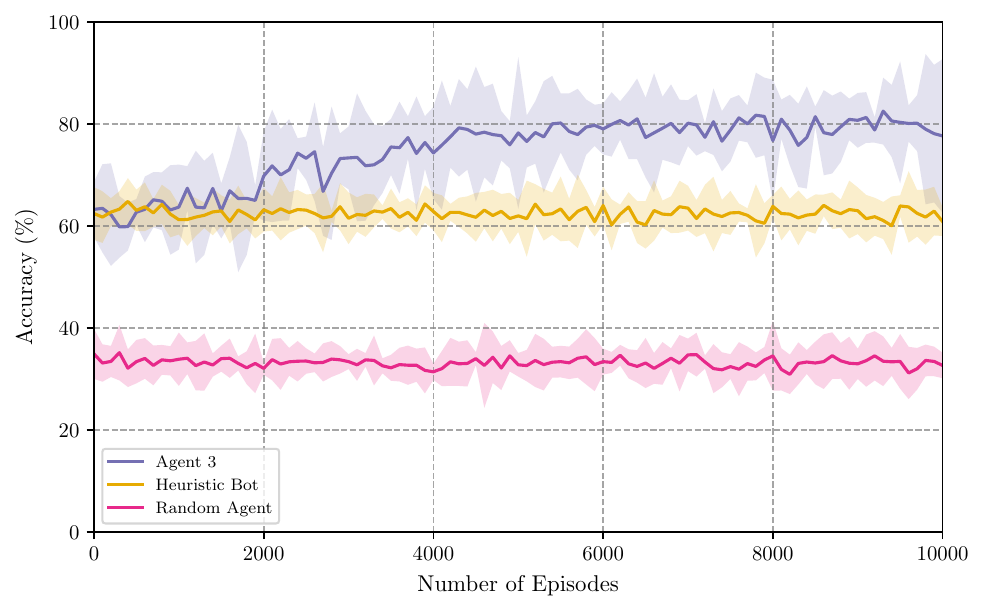}}\label{fig: dc}
  \caption{Accuracy learning curves for all 3 agents. Solid lines denote mean accuracy across all 5 respective runs. Shaded regions denote $\pm$ two standard deviations. After training for 10,000 epochs, Agents 1, 2 and 3 reach an average accuracy of 79.6\%, 80.5\% and 77.7\% respectively with all RL agents outperforming the heuristic bot.}
  \label{fig: d}
\end{figure}
 
\clearpage

\section{Appendix E - Individual Learning Curves - Relative Optimality Gap}
\label{app: e}

\begin{figure}[h]
  \centering
  \subfloat{\includegraphics[width=0.665\linewidth]{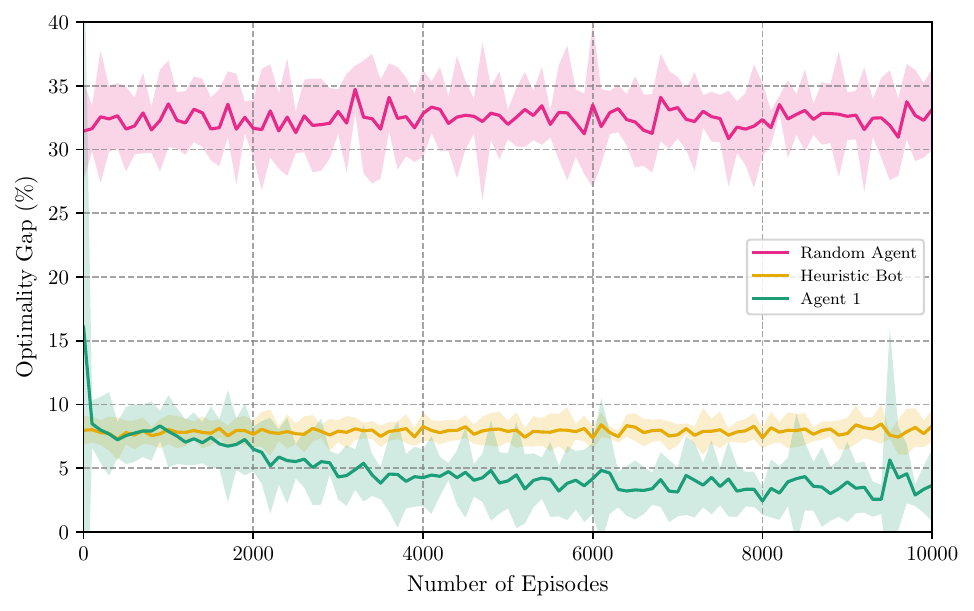}}\label{fig: e5a}
  \subfloat{\includegraphics[width=0.665\linewidth]{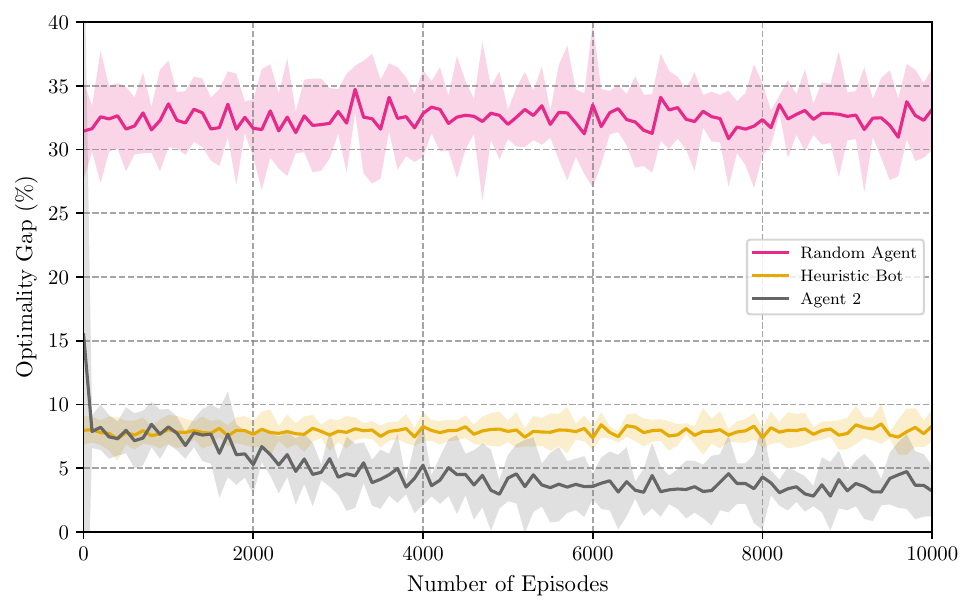}}\label{fig: e5b}
  \subfloat{\includegraphics[width=0.665\linewidth]{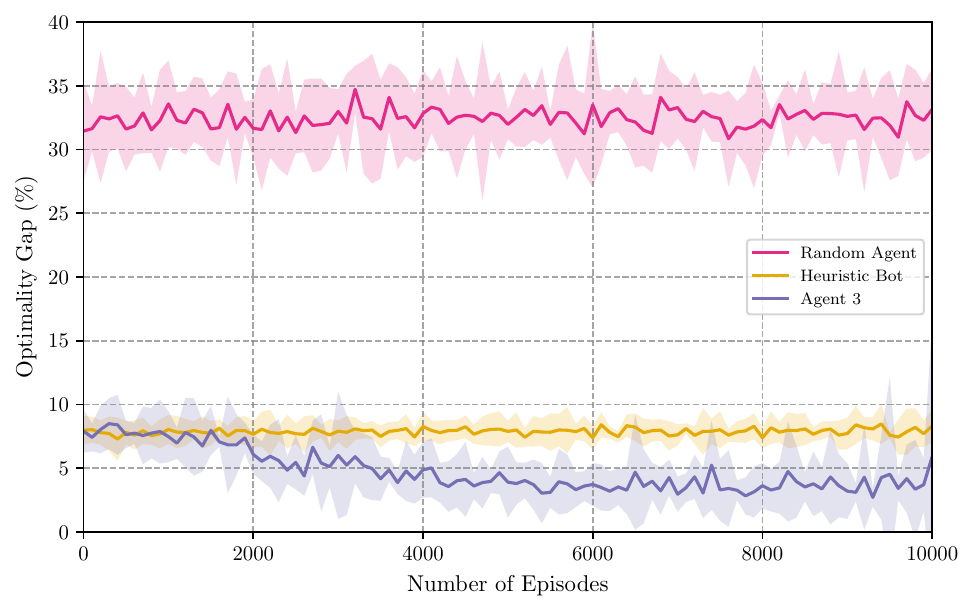}}\label{fig: e5c}
  \caption{Relative optimality gap learning curves for all 3 agents. Solid lines denote mean accuracy across all 5 respective runs. Shaded regions denote $\pm$ two standard deviations. After training for 10,000 epochs, Agents 1, 2 and 3 reach an average relative optimality gap of 3.6\%, 3.2\% and 5.9\%, with all RL agents outperforming the heuristic bot.}
  \label{fig: e}
\end{figure}

\clearpage

\section{Appendix F - Individual Learning Curves - Absolute Optimality Gap}
\label{app: f}

\begin{figure}[h]
  \centering
  \subfloat{\includegraphics[width=0.665\linewidth]{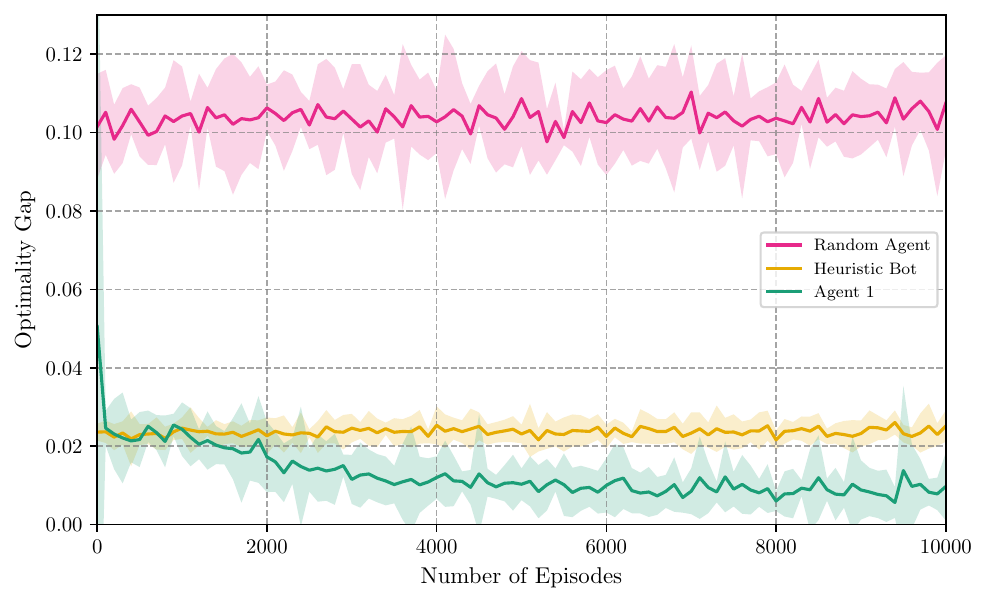}}\label{fig: f6a}
  \subfloat{\includegraphics[width=0.665\linewidth]{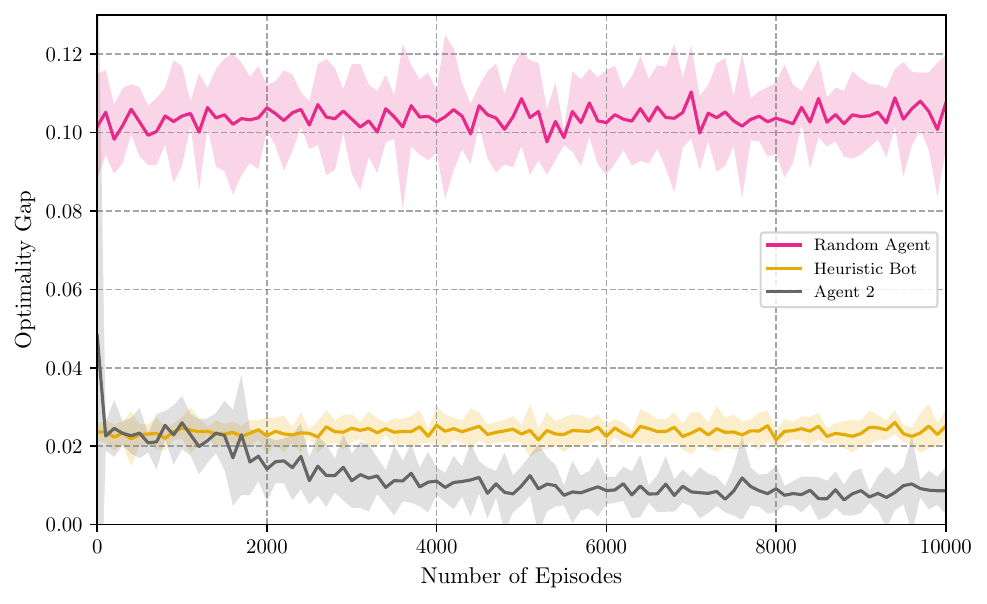}}\label{fig: f6b}
  \subfloat{\includegraphics[width=0.665\linewidth]{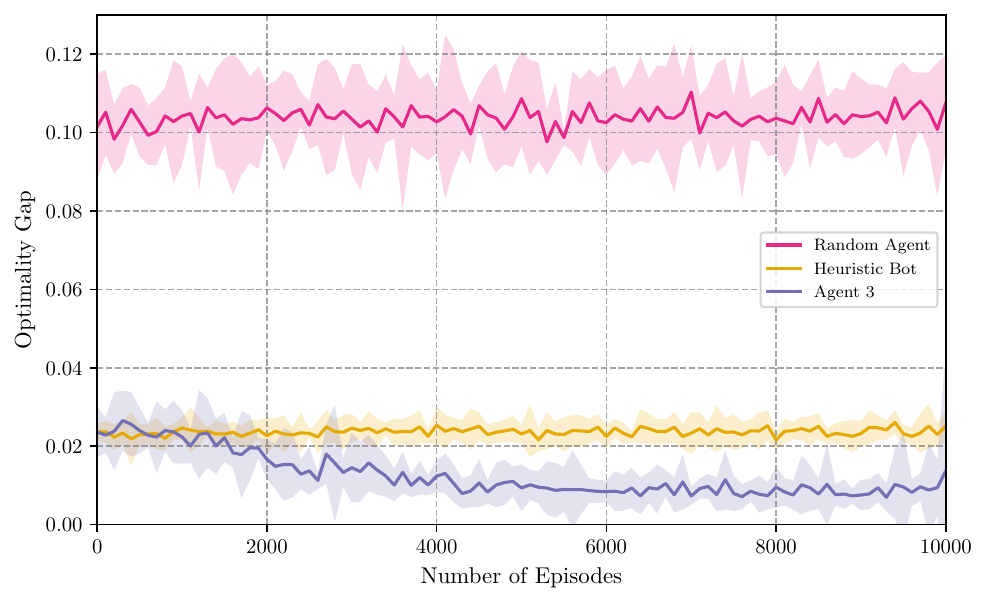}}\label{fig: f6c}
  \caption{Absolute optimality gap learning curves for all 3 agents. Solid lines denote mean accuracy across all 5 respective runs. Shaded regions denote $\pm$ two standard deviations. After training for 10,000 epochs, Agents 1, 2 and 3 reach an average absolute optimality gap of 0.010, 0.009, 0.014 with all RL agents outperforming the heuristic bot.}
  \label{fig: f}
\end{figure}

\end{document}